# Efficient shallow learning as an alternative to deep learning

Yuval Meir[1], Ofek Tevet[1], Yarden Tzach[1], Shiri Hodassman[1], Ronit D. Gross[1] and Ido Kanter[1,2*]

[1]Department of Physics, Bar-Ilan University, Ramat-Gan, 52900, Israel.

[2]Gonda Interdisciplinary Brain Research Center, Bar-Ilan University, Ramat-Gan, 52900, Israel.

[*]Corresponding author email: ido.kanter@biu.ac.il

**The realization of complex classification tasks requires training of deep learning (DL) architectures consisting of tens or even hundreds of convolutional and fully connected hidden layers, which is far from the reality of the human brain. According to the DL rationale, the first convolutional layer reveals localized patterns in the input and large-scale patterns in the following layers, until it reliably characterizes a class of inputs. Here, we demonstrate that with a fixed ratio between the depths of the first and second convolutional layers, the error rates of the generalized shallow LeNet architecture, consisting of only five layers, decay as a power law with the number of filters in the first convolutional layer. The extrapolation of this power law indicates that the generalized LeNet can achieve small error rates that were previously obtained for the CIFAR-10 database using DL architectures. A power law with a similar exponent also characterizes the generalized VGG-16 architecture. However, this results in a significantly increased number of operations required to achieve a given error rate with respect to LeNet. This power law phenomenon governs various generalized LeNet and VGG-16 architectures, hinting at its universal behavior and suggesting a quantitative hierarchical time-space complexity among machine learning architectures. Additionally, the conservation law along the convolutional layers, which is the square-root of their size times their depth, is found to asymptotically minimize error rates. The efficient shallow learning that is demonstrated in this study calls for further quantitative examination**

**using various databases and architectures and its accelerated implementation using future dedicated hardware developments.**

Traditionally, artificial neural networks have been derived from brain dynamics, where synaptic plasticity modifies the connection strength between two neurons in response to their relative activities[1,2]. The earliest artificial neural network was the Perceptron[3,4], which was introduced approximately 65 years ago, consisting of a feedforward classifier with many inputs and a single Boolean output unit. The development of more structured feedforward architectures with numerous convolutional and fully connected hidden layers, which can be increased to hundreds[5,6], as well as the development of their non-local training techniques, such as backpropagation (BP)[7,8], are required to address solutions to complex and practical classification tasks. These are essential components of the current implementation of deep learning (DL) algorithms. The underlying rationality of DL algorithms is that the first convolutional layer is sensitive to the appearance of a given pattern or symmetry in limited areas of the input, whereas the subsequent convolutional layers are expected to reveal large-scale features characterizing a class of inputs[9,10].

In a supervised learning scenario, a feedforward step is initially performed, in which the distance between the current and desired outputs for a given input is computed using a given error function. The BP procedure is utilized in the next step, where weights are updated to locally minimize the error function[7,11]. Graphic processing units (GPUs) are used to accelerate this time-consuming computational process of multiplying large matrices and vectors, and its use is repeated several times over the training set until a possible desired test error is achieved. Architectures with an increasing number of hidden layers enable learning to be efficiently optimized for complex classification tasks, which goes together with the advancement of powerful GPU technology.

However, the brain's architecture differs significantly from that of DL and consists of very few feedforward layers[12-14], only one of which approximates the convolutional wiring, mainly from the retinal input to the first hidden layer[12,15]. The key question driving our research is whether learning non-trivial classification tasks using brain-inspired shallow feedforward networks can achieve the same error rates as DL, while potentially requiring less computational complexity. A positive answer will question the need for DL architectures and might direct the development of unique hardware for the efficient and fast implementation of shallow learning. Additionally, it will demonstrate how brain-inspired shallow learning has advanced computational capability with reduced complexity and energy consumption[16,17].

**RESULTS**

LeNet[18,19], a five-layer prototype of a shallow feedforward architecture, has two convolutional layers with max-pooling operations and three successive fully connected layers (Fig. 1A). The first and second convolutional layers have $d_1 = 6$ and $d_2 = 16$ filters, respectively, representing the depth of each layer, and their convolutional layer sizes after max-pooling, $m_i \times m_i$ $(height \times width)$, are $14 \times 14$ and $5 \times 5$, respectively. One can notice that

$$\frac{d_2}{d_1} = \frac{16}{6} \simeq \frac{14}{5} = \frac{m_1}{m_2} = 2.8 \quad (1)$$

which hints on the following conservation law along the convolutional layers

$$depth_i \times m_i = constant \quad (2)$$

where $m_i^2$ and $depth_i$ represent the $i^{th}$ convolutional layer size and the number of filters, respectively. We minimized the LeNet error rates for the CIFAR-10 database[20] as a function of $d_1$ while maintaining the ratio $d_2/d_1$ constant (Fig. 1B and Supplementary Information). The results indicate decaying of error rates, $\epsilon$, with increasing $d_1$ as a power law

$$\epsilon(d_1) = \frac{A}{(d_1)^\rho} \quad (3)$$

with an exponent $\rho \sim 0.4$, even for small $d_1$. Although the error rate of the original LeNet, $d_1 = 6$, is $\epsilon \simeq 0.24$, it can be further minimized by increasing $d_1$. According to the extrapolation of the power law to large $d_1$ values, any small $\epsilon$ can be achieved using LeNet, a shallow architecture. However, its minimization for a given large $d_1$ is a heavy computational task that requires an exhaustive search in a hyper-parameter space that its values vary among layers, with an increasing number of epochs and complex scheduling. For instance, preliminary results of an incomplete optimization for $d_1 = 27$ and $d_2 = 72$ using at least 500 epochs indicate $\epsilon \sim 0.137$, which is close to the expected result of the extrapolated power law (Fig. 1B).

The conservation law (Eq. (2)) was found to govern the convolutional layer sizes of the original VGG-16 architecture, which consists of 16 layers[21], except for the fifth convolution set, where the number of filters is bounded by 512 (Fig. 2A with $d = 64$). The $n^{th}$ ($n \leq 4$) convolution set has $d \cdot 2^{n-1}$ filters, where the convolutional layer size is $\frac{m}{2^{n-1}}$ ($n \leq 5$). The minimization of $\epsilon$ for VGG-16 and the CIFAR-10 database (Fig. 2A with $m = 32$) as a function of $d$ results in a power law with a similar exponent to LeNet (Fig. 1B), $\rho \sim 0.4$ (Fig. 2B and Supplementary Information). The results allude to the universal behavior of power-law scaling (Eq. (3)) which is independent of the architecture details, where $d$ is the number of filters in the first convolutional layer. Additionally, the exponent, $\rho$, does not necessarily increase with the number of convolutional or hidden layers. Interestingly, the standard VGG-16 network ($d = 64$ in Fig. 2A), with batch normalization but without dropouts, results in $\epsilon \sim 0.065$ (Supplementary Information), which is identical to the reported test error with significant dropouts[22]. Hence, the advantage of dropouts in the minimization of $\epsilon$ might be questionable in this case.

A shallow network's ability to achieve any small $\epsilon$, based on the extrapolation of the power-law scaling (Fig. 1B), is accompanied by a significant reduction in computational complexity per epoch compared with a DL architecture (Fig. 2A). Complexity is measured as the number of multiplication-add (MAdd) operations per input during a forward and BP step[23,24]. It is calculated as a function of the number of filters, $d_1$ and $d$ in Figs. 1 and 2, respectively (Fig. 3A, Supplementary Information). In both cases, the number of operations per step scale as a quadratic polynomial with the number of filters and are derived from the following argument: When the number of filters in a convolutional layer is doubled, its computational complexity increases by a factor of four because its consecutive convolutional layer is also doubled. The origin of the linear terms in the quadratic polynomials (Fig. 3A) is mainly attributed to the input size of the first fully connected layer, which increased linearly with the number of filters. Hence, the number of weights increases linearly with the number of filters, whereas the number of weights in the successive fully connected layers remains constant and is independent of the number of filters (Figs. 1A and 2A).

The computational complexities as a function of the error rates were calculated using the power law extrapolation of $\epsilon(d)$ (Figs. 1B and 2B). The results indicate that the complexity increases with $1/\epsilon$, as a power law with an exponent $\rho$ close to 5, $\sim 4.95$ for LeNet and $\sim 4.94$ for VGG-16 (Fig. 3B). Since error rates in both cases (Figs. 1B and 2B) are approximated by

$$\epsilon \propto \frac{1}{d^{0.4}},$$

therefore, $d \propto \epsilon^{-2.5}$ and in the leading term (Fig. 3A)

$$Complexity \propto d^2 \propto \epsilon^{-5}$$

A direct calculation of the computational complexity ratio per step between LeNet and VGG-16, based on $\epsilon(d)$ (Figs. 1B and 2B), indicates that it is less than 0.7 for at least $\epsilon \geq 0.005$ (Fig. 3C). As it is extremely sensitive to the similar estimated values of $\rho$ for both LeNet and VGG-16 (Figs. 1B and 2B), further extrapolation toward vanishing $\epsilon$ is unclear. Nevertheless, the lower complexity per epoch of shallow architectures serves as an example of the potential advantages of brain-inspired architectures. We note that the entire computational learning complexity is proportional to the number of training epochs and the classification of an input depends on a forward step only.

Under parallel computation, the required number of clock steps in a feedforward or BP realization is bounded from below by the number of layers. Decreasing this lower bound using a mechanism similar to that of carry-lookahead[25], developed for the addition and multiplication of large numbers, is practically inapplicable for such complex architectures. This is another expected advantage of learning based on brain-inspired shallow architecture.

The power-law behavior (Eq. (3)) is demonstrated to govern both shallow and DL architectures, where the number of filters obeys the conservation law (Eq. (2)). The following two questions were examined: The first question concerns the robustness of the power law (Eq. (3)) for architectures that deviate from the conservation law (Eq. (2)). The second question is whether Eq. (2), which controls the number of filters in the convolutional layers is indeed the optimized choice to minimize $\epsilon$.

The power-law scaling for LeNet, which deviates from the conservation law (Eq. (2)) is defined as follows:

$$\frac{d_2}{d_1} = constant \neq \frac{16}{6} \quad (4)$$

For a smaller constant, $\frac{4}{3}$, the error rates were increased by a larger pre-factor $A$, as shown in Eq. (3); however, $\rho$ remained similar $\sim 0.4$ (Fig. 4A). For a larger constant, $\frac{16}{3}$, the slope decreased, $\rho \sim 0.35$ (Fig. 4A). The results first indicate the robustness of the power law for various constants (Eq. (4)) which alludes to its universal behavior. Second, for a smaller constant and any given $d_1$, the error rates were enhanced. For a large constant (Eq. (4)) and sufficiently large $d_1$ error rates were also enhanced because $\rho$ decreased, but for small $d_1$ values, the error rates decreased. The results indicate that the conservation law (Eq. (2)) with a constant that is expected to be approximately $\frac{16}{6}$, asymptotically minimizes $\epsilon$ for a large $d_1$. Similar trends were obtained for VGG-16, where the number of filters in the $n^{th}$ convolution set ($n \leq 4$) increased as $constant^{(n-1)}$, whereas in the original architecture $constant = 2$ (Fig. 2A). For $constant = 1.5$, the error rates increased with a larger pre-factor, $A$, where $\rho$ remained similar $\sim 0.4$ (Fig. 4B). For $constant = 2.5$, $\rho \sim 0.32$, indicating once more that the error rates increased asymptotically compared to $constant = 2$, but for small $d$, the error rates could be decreased. The results for VGG-16 indicate the robustness of the universal power-law behavior for various constants, as shown in Eq. (4), where a $constant$ close to 2 minimizes $\epsilon$.

The following theoretical justification may explain why the conservation law (Eq. (2)) leads to the minimization of error rates: Its purpose is to preserve the signal-to-noise ratio (SNR) along the feedforward convolutional layers such that the signal is repeatedly amplified. The noise of each large convolutional filter is expected to be proportional to the square-root of its size, $m$, and its signal to $m^2$. Consequently, the SNR is proportional to $m$, and for the entire convolutional layer composed of depth $d$ is $m \cdot d$. Hence, to compensate for the shrinking of the convolutional layer size along the feedforward architecture, its depth must be increased accordingly. Indeed, preliminary results indicate that doubling the number of filters in the fifth convolution set of VGG-16 (with $d = 16$), such that the number of filters in all convolutions ($n \leq 5$) is $16 \cdot 2^{n-1}$ decreased $\epsilon$ by ~0.015 compared with the standard VGG-16 architecture (Fig. 2 and Supplementary Information). This supports the argument that maintaining the same SNR along the entire deep architecture enhances success rates. Nevertheless, further extended simulations on various architectures and databases are required to support the accuracy of the suggested conservation law, particularly because the convolutional layer sizes are small and far from the thermodynamic limit. Additionally, it is important to examine how the sensitivity of $\rho$ and the conservation law are related to the properties of the cost function and the details of BP dynamics.

**DISCUSSION**

Minimizing error rates for a particular classification task and database has been one of the primary goals of machine learning research over the past few decades. As a result, more structured DL architectures consisting of various combinations of concatenated convolutional and densely connected layers have been developed. Typically, further significant minimization of error rates requires deeper architectures, where such an architecture with some modifications can achieve reasonably high success rates for several other databases and classification tasks. This study suggests that, using the extrapolation of the power-law scaling (Eq. (3)) traditional shallow architectures can achieve the same error rates as state-of-the-art DL architectures. The preferred architecture can reduce the space–time complexity for a specific training algorithm on a given database and hardware implementation. A theoretical framework is presented for constructing a hierarchy of complexity between families of artificial feedforward neural network architectures, based on their power-law scaling, exponent $\rho$, and pre-factor $A$. It is possible that the optimal architecture among several ones depends on the desired error rate (Fig. 4). Contrary to common knowledge, shallow feedforward brain-inspired architectures are not inferior, and they do not represent, as thought, an additional biological limitation[26]. They can achieve low error rates such as DL algorithms, even with significantly low computational complexity for complex classification tasks (Fig. 3).

Architectures that maximize $\rho$ and its upper bound are not yet known. Preliminary results indicate that for a specific architecture, $\rho$ may increase when the number of weights grows super-linearly with the number of filters. This can be achieved using a fully connected layer, in which the number of input and output units is proportional to the number of filters. Another possible mechanism is the addition of a super-linear number of cross-weights to the filters. This represents a biological realization because cross-weights result as a byproduct of dendritic nonlinear amplification[17,27-29]. Nevertheless, these possible enhanced $\rho$ mechanisms significantly increase computational complexity and are mentioned for their potential biological relevance, limited number of layers, and the natural emergence of many cross-weights.

Advanced GPU technology is used to minimize the running time of the DL algorithms. Indeed, our single-epoch running time using the CIFAR-10 database and VGG-16 with $d = 4$ is only a factor ~1.5 compared with LeNet with $d_1 = 6$, where both cases have similar success rates. However, shallow architectures with the same error rates as advanced deep architectures require more filters per convolutional layer, and consequently, a significantly increased number of fully connected weights. Above a critical number of filters, depending on the GPU properties, an epoch's running time is significantly slowed down and can even increase by a few orders of magnitude. Similarly, the running time in our case of VGG-16 with $d = 400$ is ~60 times slower than that with $d = 8$, and LeNet with $d_1 = 2304$ is ~900 times slower than that with $d_1 = 6$. Hence, efficient realization of competitive error rates of shallow architectures to advance DL architectures requires a shift in the properties of advanced GPU technology. Additionally, it is expected to achieve a significant reduction in computational complexity for a desired error rate and a specific database (Fig. 3).

Finally, the theoretical origin of the universal power-law scaling (Eq. (3)), governing shallow and DL architectures, has not yet been discovered. The following theoretical framework may provide a starting point for investigating this general phenomenon. The teacher-student online scenario is one of the analytically solvable cases exemplifying power law behavior cases[30]. In the prototypical realizable scenario, the teacher and student have the same feedforward architecture, for example, a binary or a soft committee machine[31-33], but different initial weights. The teacher supports the students with a random input-output relation, and the student updates its weights based on this information and its current set of weights. The generalization error (test error) decays as a power law with the number of input-output examples, which is normalized to the size of the input. This work differs from online learning because the size of the non-random training dataset is limited, and a training example is repeatedly presented as an input without an online scenario. However, assuming a power-law scaling (Eq. (3)), an architecture with an infinite number of filters $d \to \infty$, exists such that the test error vanishes. This architecture is the teacher's counterpart and represents a learning rule in the online scenario. A student with fewer filters attempts to imitate the teacher and results in a generalization error, which is expected to decrease with an increasing number of filters. It is currently impossible to find an analytical solution for the shallow and deep architectures that are shown as a function of the number of filters. The question is whether a toy model, where a filter may be represented by a perceptron with a nonlinear output unit, can be solved analytically to show that the generalization error decays as a power law with the number of filters.

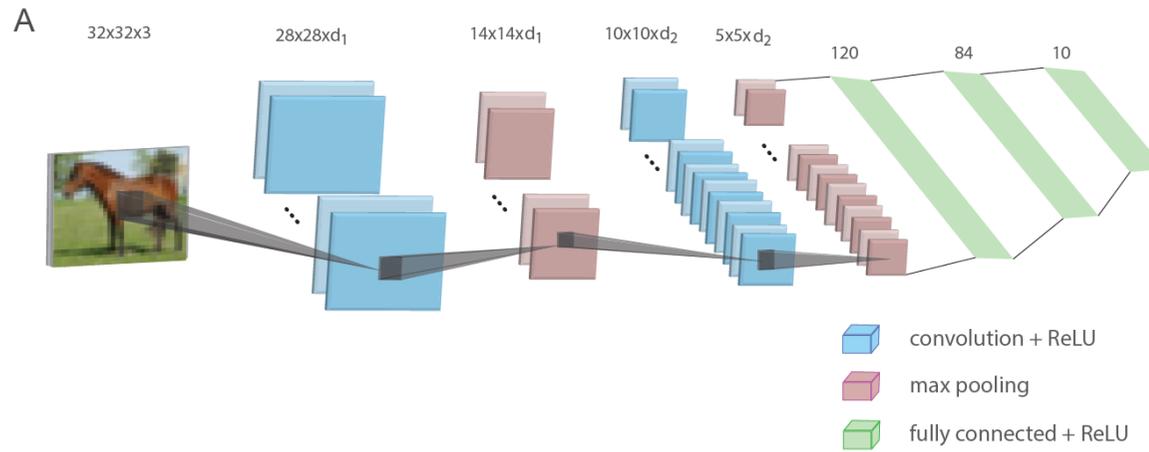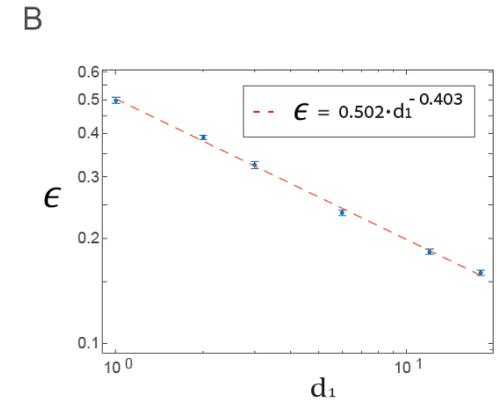

**Figure 1. Learning in generalized LeNet architecture. (A)** Generalized LeNet architecture for the CIFAR-10 database (32 × 32 × 3 pixels input's size) consisting of five layers, two convolutional layers including max pooling and three fully connected layers. The first and second convolutional layers consist of $d_1$ and $d_2$ filters, respectively, where $\frac{d_1}{d_2} \simeq \frac{6}{16}$. **(B)** The test error, $\epsilon$, as a function of $d_1$ on a log-log scale, indicating a power-law scaling with exponent $\rho \sim 0.4$, Eq. (3) (Supplementary Information). The activation function of the nodes is ReLU.

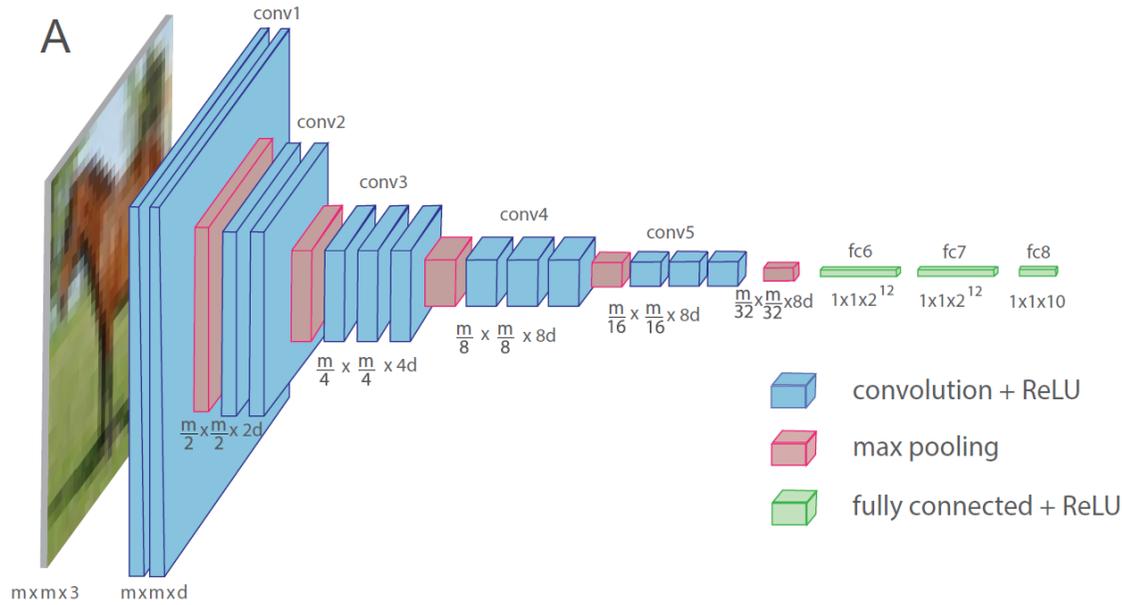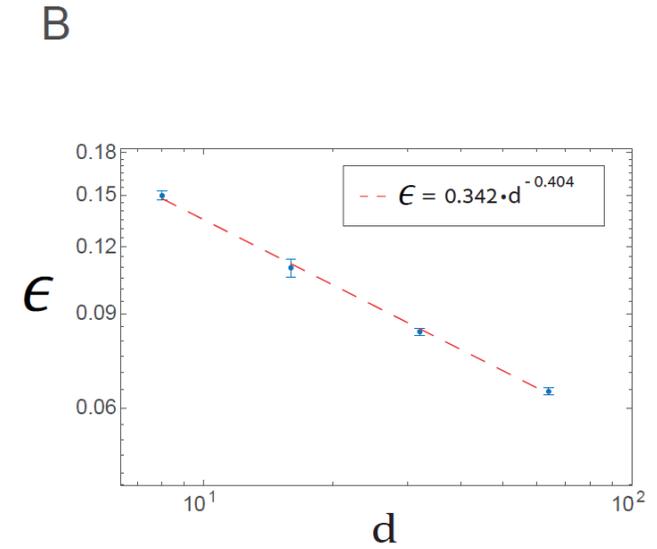

**Figure 2. Learning in generalized VGG-16 architecture.** **(A)** Generalized VGG-16 architecture consisting of 16 layers, where the number of filters in the $n^{th}$ convolution set is $d \cdot 2^{n-1}$ ($n \leq 4$) and the square-root of the size of the filter is $m \cdot 2^{-(n-1)}$ ($n \leq 5$), where $m \times m \times 3$ is the size of each input ($d = 64$ in the original VGG-16 architecture). **(B)** The test error, $\epsilon$, as a function of $d$ on a log-log scale, for the CIFAR-10 database ($m = 32$), indicating a power-law scaling with exponent $\rho \sim 0.4$, Eq. (3) (Supplementary Information). The activation function of the nodes is ReLU.

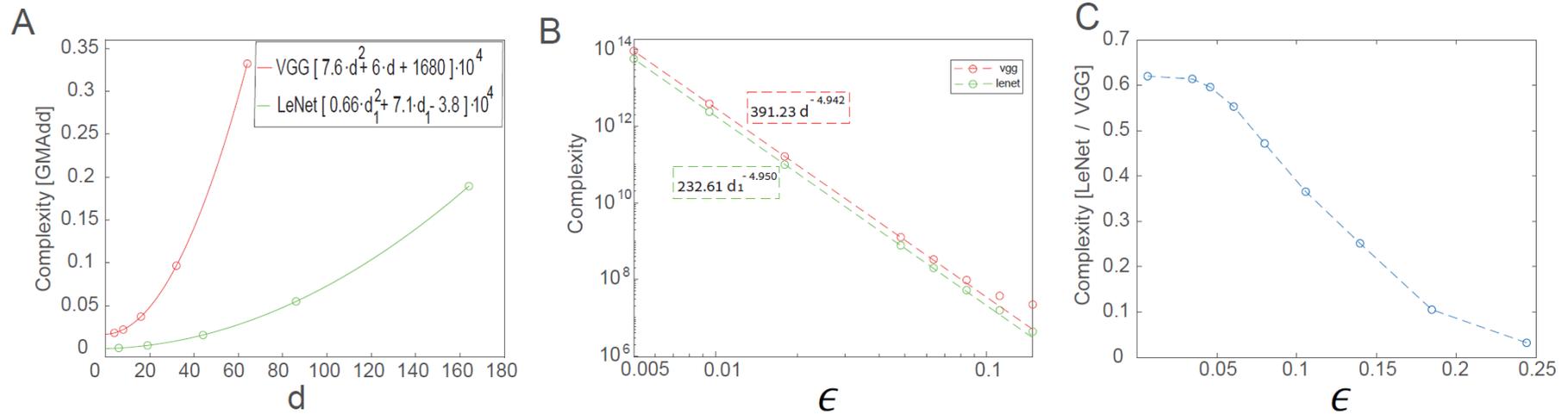

**Figure 3. A comparison of learning complexity between generalized LeNet and VGG-16 architectures. (A)** Complexity of a feedforward and BP step for a single input of LeNet (green) and VGG-16 (red) measured for several $d$ ($d_1$) values (open circles) and the quadratic polynomial fits. Complexity is measured in Giga multiplication-add operations (GMAdd). **(B)** Complexity as a function of $\epsilon(d)$ for LeNet (green) and VGG-16 (red) for several values of small $\epsilon$ (open circles) obtained from the power-law scaling (Figs. 1B and 2B) and the fitted power-law scaling (in dashed boxes), obtained from the last three small values of $\epsilon$ (dashed lines). **(C)** The ratio between the complexity of LeNet and VGG-16 for several values $\epsilon$ (open circles connected by a dashed line), obtained from the extrapolated $\epsilon(d_1)$ and $\epsilon(d)$ for LeNet and VGG-16 (Figs. 1B and 2B), respectively, and a direct measure of the complexity[23,24] (Supplementary Information).

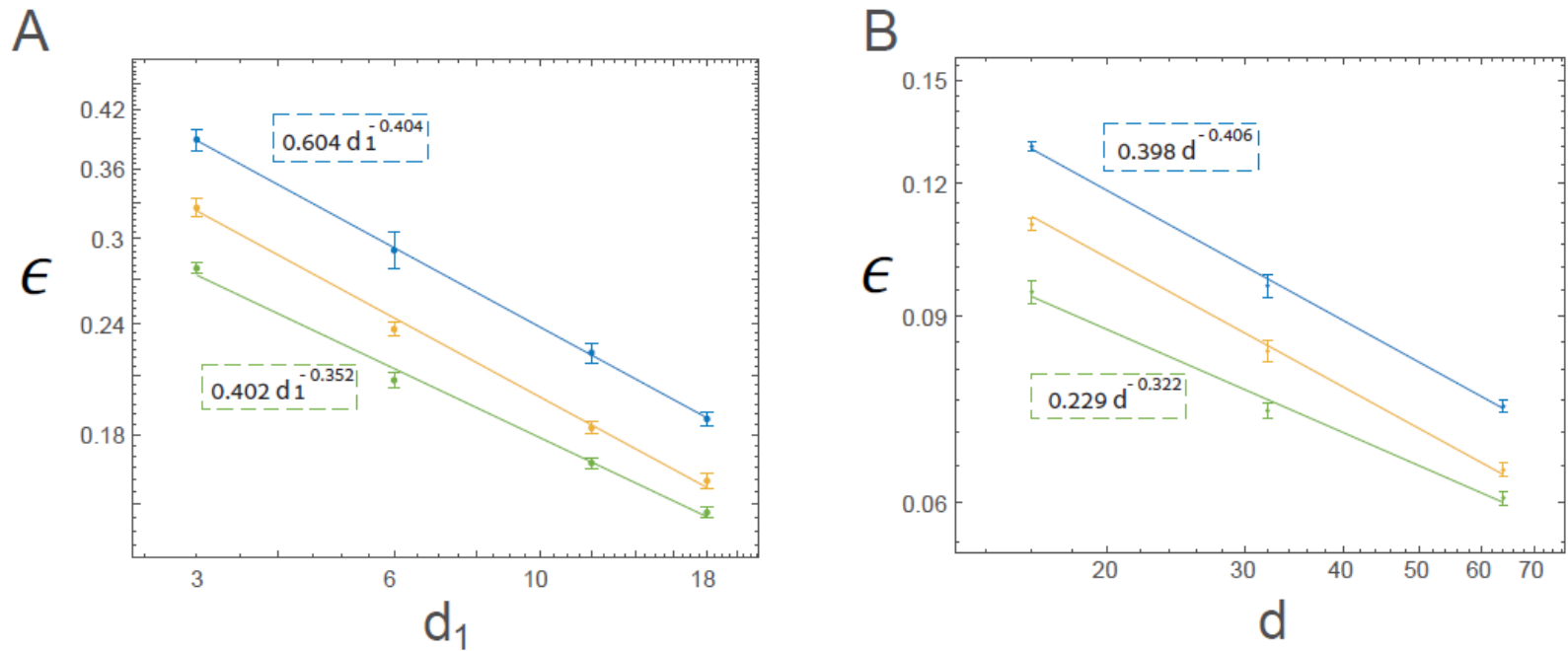

**Figure 4. Conservation law indicating the optimal ratio between the depth of filters and their convolutional layer size. (A)** Success rates and their standard deviation as a function of $d_1$ for the generalized LeNet architecture where $\frac{d_2}{d_1} = \frac{16}{3}$ (green) and $\frac{d_2}{d_1} = \frac{4}{3}$ (blue) with fitted (color coded) power law (Supplementary Information), and the results for $\frac{d_2}{d_1} = \frac{16}{6}$ (yellow, as shown in Fig. 1B) are presented for reference. **(B)** Similar to **A**, for the generalized VGG-16 architecture, where the number of filters in the $n^{th}$ convolution set ($n \leq 4$) is $d \cdot 2.5^{n-1}$ (green) and $d \cdot 1.5^{n-1}$ (blue) with fitted (color coded) power law (Supplementary Information), and results for $d \cdot 2^{n-1}$ (yellow, as shown in Fig. 2B) are presented for a reference.

**Data availability**

Source data are provided in this study, including all data supporting the plots, along with other findings of this study.

**Acknowledgments**

I.K. acknowledges the partial financial support from the Israel Science Foundation (grant number 346/22). S.H. acknowledges support from the Israeli Ministry of Science and Technology. We thank Amir Goldental and Roni Vardi for their stimulating discussions.

**Author contributions**

Y.M. contributed considerably to all simulations on LeNet and VGG-16, O.T. derived the main results for LeNet, R.G. contributed considerably to LeNet simulations, Y.T. contributed considerably to all simulations, and S.H. contributed to simulations and the preparation of the figures. I.K. initiated the study and supervised all aspects of the study. All authors have commented on the manuscript.

**Competing interests**

The authors declare no competing interests.

# Supplementary Information

# Efficient shallow learning as an alternative to deep learning

**Yuval Meir[1], Ofek Tevet[1], Yarden Tzach[1], Shiri Hodassman[1], Ronit D. Gross[1] and Ido Kanter[1,2*]**

[1]Department of Physics, Bar-Ilan University, Ramat-Gan, 52900, Israel.

[2]Gonda Interdisciplinary Brain Research Center, Bar-Ilan University, Ramat-Gan, 52900, Israel.

[*]Corresponding author email: ido.kanter@biu.ac.il

**Generalized LeNet and VGG-16 architectures and initial weights.** The generalized LeNet architecture (Fig. 1A) consists of two consecutive convolutional layers of size $(5 \times 5)$ and depths $d_1$ and $d_2$ and three fully connected layers of sizes $25 \times d_2, 120, 84$. After each convolutional layer, max-pooling consisting of $(2 \times 2)$ was applied[1]. The softmax function was applied to the ten outputs. The generalized VGG-16 architecture (Fig. 2A) consists of 16 layers[2], 13 convolutional layers with a single zero-padding around each input, and three fully connected layers with 4096 hidden units each. In the first four convolutional layers, max-pooling consisting of $(2 \times 2)$ was applied every two convolutions. In the rest of the nine convolutional layers, a max-pooling layer was applied after every three convolutional layers. After each convolutional layer, a batch normalization layer was applied. The softmax function was applied to the ten outputs. In both architectures, the ReLU activation function was assigned to each hidden unit, and all weights were initialized using a Gaussian distribution with a zero mean and standard deviation according to the He normal initialization[3].

**Data preprocessing.** Each input pixel of an image $(32 \times 32)$ from the CIFAR-10 database was divided by the maximal pixel value, 255, multiplied by 2, and subtracted by 1, such that its range was $[-1, 1]$. In all simulations, data augmentation was used, derived from the original images, by random horizontally flipping and translating up to four pixels in each direction.

**Optimization.** The cross-entropy cost function was chosen for the classification task and was minimized using the stochastic gradient descent algorithm[4,5]. The maximal accuracy was determined by searching through the hyper-parameters (see below). Cross-validation was confirmed using several validation databases, each consisting of 10,000 random examples, as in the test set. The averaged results were in the same standard deviation (Std) as the reported average success rates. The Nesterov momentum[6] and L2 regularization method[7] were applied.

**Hyper-parameters.** The hyper-parameters η (learning rate), μ (momentum constant[6]), and α (regularization L2[7]) were optimized for offline learning, using a mini-batch size of 100 inputs selected such that each of the ten labels of the

CIFAR-10 dataset appear ten times. The learning rate decay schedule[8,9] was also optimized such that it was multiplied by the decay factor, q, every Δt epochs, and was denoted as (q, Δt).

| Figure 1 | | | | | |
|---|---|---|---|---|---|
| $d_1$ | $d_2$ | η | μ | α | epochs |
| 1 | 2 and 3 | 0.028 | 0.850 | 9.5e-4 | 240 |
| 2 | 5 and 6 | 0.028 | 0.850 | 9.5e-4 | 240 |
| 3 | 8 | 0.028 | 0.905 | 9.5e-4 | 220 |
| 6 | 16 | 0.028 | 0.910 | 9.5e-4 | 280 |
| 12 | 32 | 0.028 | 0.915 | 9.5e-4 | 240 |
| 18 | 48 | 0.028 | 0.950 | 9.5e-4 | 280 |

The decay schedule for the learning rate is defined as follows:

$$(q, \Delta t) = \begin{cases} (0.8, 10) & \text{epoch} \leq 120 \\ (0.7, 10) & \text{epoch} > 120 \end{cases}$$

For LeNet, the conservation constant was $\frac{d_2}{d_1} = \frac{16}{6} = \frac{8}{3}$. For $d_1 = 1$ and $d_1 = 2$ the conservation law results in non-integer values $d_2 = 2\frac{2}{3}$ and $d_2 = 5\frac{1}{3}$, respectively. Assuming a power law, the following interpolation was performed for $d_1 = 1$:

$$\ln\left(\epsilon(d_2 = 2\frac{2}{3})\right) = \frac{\ln\left(2\frac{2}{3}\right) - \ln(2)}{\ln(3) - \ln(2)} \cdot \ln(\epsilon(d_2 = 2)) + \frac{\ln(3) - \ln\left(2\frac{2}{3}\right)}{\ln(3) - \ln(2)} \cdot \ln(\epsilon(d_2 = 3))$$

and for $d_1 = 2$:

$$\ln\left(\epsilon(d_2 = 5\frac{1}{3})\right) = \frac{\ln\left(5\frac{1}{3}\right) - \ln(5)}{\ln(6) - \ln(5)} \cdot \ln(\epsilon(d_2 = 5)) + \frac{\ln(6) - \ln\left(5\frac{1}{3}\right)}{\ln(6) - \ln(5)} \cdot \ln(\epsilon(d_2 = 6))$$

Note that the weighted arithmetic mean yields the same results (at least up to two leading digits).

| Figure 2 | | | | |
|---|---|---|---|---|
| d | η | μ | α | epochs |
| 8 | 0.01 | 0.920 | 9e-4 | 200 |
| 16 | 0.01 | 0.975 | 1.5e-3 | 200 |
| 32 | 0.01 | 0.965 | 9.5e-4 | 200 |
| 64 | 0.028 | 0.975 | 1.5e-3 | 200 |

The decay schedule for the learning rate is defined as follows:

$$(q, \Delta t) = (0.6, 20)$$

| Figure 4A | | | | | |
|---|---|---|---|---|---|
| Constant = $\frac{4}{3}$ | | | | | |
| $d_1$ | $d_2$ | $\eta$ | $\mu$ | $\alpha$ | epochs |
| 3 | 4 | 0.035 | 0.900 | 1e-5 | 200 |
| 6 | 8 | 0.030 | 0.975 | 1e-5 | 200 |
| 12 | 16 | 0.030 | 0.965 | 4e-5 | 200 |
| 18 | 24 | 0.025 | 0.975 | 2e-4 | 200 |

The decay schedule for the learning rate for $d_1 = 3, 6$ and 12 is defined as follows:

$$(q, \Delta t) = \begin{cases} (0.9, 10) & \text{epoch} \leq 60 \\ (0.85, 10) & \text{epoch} > 60 \end{cases}$$

For $d_1 = 18$, the schedule is defined as follows:

$$(q, \Delta t) = \begin{cases} (0.95, 10) & \text{epoch} \leq 30 \\ (0.9, 10) & 30 < \text{epoch} \leq 60 \\ (0.8, 10) & 60 < \text{epoch} \leq 200 \end{cases}$$

| Constant = $\frac{16}{3}$ | | | | | |
|---|---|---|---|---|---|
| $d_1$ | $d_2$ | $\eta$ | $\mu$ | $\alpha$ | epochs |
| 3 | 16 | 0.028 | 0.940 | 9e-4 | 200 |
| 6 | 32 | 0.006 | 0.975 | 9e-4 | 200 |
| 12 | 64 | 0.010 | 0.975 | 9e-4 | 200 |
| 18 | 96 | 0.010 | 0.975 | 1.5e-3 | 200 |

The decay schedule for the learning rate for $d_1 = 3$ is defined as follows:

$$(q, \Delta t) = \begin{cases} (0.8, 10) & \text{epoch} \leq 120 \\ (0.7, 10) & \text{epoch} > 120 \end{cases}$$

For $d_1 = 6, 12$, and 18, the schedule is defined as follows:

$$(q, \Delta t) = (0.6, 20)$$

| Figure 4B |   |   |   |   |
|---|---|---|---|---|
| Constant = $\frac{3}{2}$ | | | | |
| d | η | μ | α | epochs |
| 16 | 0.008 | 0.975 | 9e-4 | 200 |
| 32 | 0.007 | 0.975 | 1.5e-3 | 200 |
| 64 | 0.002 | 0.970 | 3e-3 | 200 |
| Constant = $\frac{5}{2}$ | | | | |
| d | η | μ | α | epochs |
| 16 | 0.010 | 0.975 | 9e-4 | 200 |
| 32 | 0.010 | 0.965 | 9e-4 | 200 |
| 64 | 0.015 | 0.975 | 9e-4 | 200 |

The decay schedule for the learning rate is defined as follows:

$$(q, \Delta t) = (0.6, 20)$$

**Generalization error** ($\epsilon$)**.** The evaluation of $\epsilon$ is performed by testing the network on 10000 test examples, that were not used in the training. The network's decision was taken as the output unit with the maximum value among the ten output units.

**Raw data of the graphs.** The raw data for the four graphs are presented in the following tables.

| Figure 1 | | | |
|---|---|---|---|
| $d_1$ | $d_2$ | $\epsilon$ | Std |
| 1 | 2 | 0.534 | 0.0205 |
| 1 | 3 | 0.481 | 0.0072 |
| 1 | 2 and 3 | 0.498 | 0.0097 |
| 2 | 5 | 0.396 | 0.0061 |
| 2 | 6 | 0.376 | 0.0045 |
| 2 | 5 and 6 | 0.389 | 0.0054 |
| 3 | 8 | 0.325 | 0.0077 |
| 6 | 16 | 0.236 | 0.0045 |
| 12 | 32 | 0.183 | 0.0029 |
| 18 | 48 | 0.159 | 0.0031 |

| Figure 2 | | |
|---|---|---|
| d | $\epsilon$ | Std |
| 8 | 0.1496 | 0.0027 |
| 16 | 0.1097 | 0.0040 |
| 32 | 0.08334 | 0.0011 |
| 64 | 0.0644 | 0.0019 |

| Figure 3A (LeNet) | |
|---|---|
| $d_1$ | MAdds |
| 6 | 651.72K |
| 19 | 3.7M |
| 44 | 15.82M |
| 86 | 54.99M |
| 164 | 190M |

| Figure 3A (VGG-16) | |
|---|---|
| d | MAdds |
| 4 | 18.28M |
| 8 | 22.17M |
| 16 | 37.25M |
| 32 | 96.61M |
| 64 | 332M |

| Figure 3B | | |
|---|---|---|
| $\epsilon$ | GMAdd (LeNet) | GMAdd (VGG-16) |
| 0.0481 | 0.77 | 1.27 |
| 0.0180 | 100 | 163 |
| 0.0095 | 2380 | 3860 |
| 0.0050 | 57520 | 92480 |

| Figure 3C | |
|---|---|
| $\epsilon$ | Complexity ratio [LeNet / VGG-16] |
| 0.0637 | 0.600 |
| 0.0481 | 0.610 |
| 0.0180 | 0.613 |
| 0.0095 | 0.616 |
| 0.0050 | 0.622 |

| Figure 4A | | |
|---|---|---|
| Constant = $\frac{4}{3}$ | | |
| $d_1$ | $\epsilon$ | Std |
| 3 | 0.388 | 0.0108 |
| 6 | 0.291 | 0.0140 |
| 12 | 0.222 | 0.0058 |
| 18 | 0.187 | 0.0036 |
| Constant = $\frac{16}{3}$ | | |
| $d_1$ | $\epsilon$ | Std |
| 3 | 0.277 | 0.0029 |
| 6 | 0.207 | 0.0043 |
| 12 | 0.167 | 0.0013 |
| 18 | 0.146 | 0.0010 |

| Figure 4B | | |
|---|---|---|
| Constant = $\frac{3}{2}$ | | |
| d | $\epsilon$ | Std |
| 16 | 0.129 | 0.0012 |
| 32 | 0.096 | 0.0022 |
| 64 | 0.074 | 0.0009 |
| Constant = $\frac{5}{2}$ | | |
| d | $\epsilon$ | Std |
| 16 | 0.094 | 0.0022 |
| 32 | 0.073 | 0.0011 |
| 64 | 0.060 | 0.0008 |

**Enhanced VGG-16.** The enhanced architecture extends the number of filters following the conservation law to the fifth convolution set (Fig. 2A). At convolutional layers 11-13 the number of filters is extended to $16d$ according to the conservation law, instead of $8d$ in the original architecture. This extended architecture was tested on an initial filter size of $d = 16$, resulting in $256$ filters for the fifth convolution set and yielding an average error rate of $0.0937$, an improvement from the original VGG-16 ($0.11$).

| Learning in enhanced VGG-16 | | | | |
|---|---|---|---|---|
| $d_1$ | η | μ | α | epochs |
| 16 | 0.007 | 0.975 | 2.0e-3 | 200 |

The decay schedule for the learning rate is defined as follows:

$$(q, t) = (0.6, 20)$$

**Multi-Adds calculations.** The complexity of each network was calculated based on the number of multiplication-add operations performed [10] in the forward and BP of a single input image.

**Statistics.** Statistics for all simulations were obtained using at least 10 samples.